\documentclass[10pt,twocolumn]{article}
\usepackage{latex8}

\usepackage{amsmath}
\usepackage{booktabs}

\usepackage{url}

\usepackage[ruled]{algorithm2e}
\usepackage[]{algorithm2e}

\usepackage{arabtex}

\usepackage{epsfig}
\usepackage{url}
\usepackage{float}
\usepackage{graphicx}

\SetAlFnt{\small}
\SetAlCapFnt{\small}
\SetAlCapNameFnt{\small}
\SetAlCapHSkip{0pt}
\IncMargin{\parindent}

\title{Contextual Analysis for Middle Eastern Languages with Hidden
  Markov Models}
\author{Kazem Taghva\\
Department of Computer Science\\
University of Nevada, Las Vegas\\
Las Vegas, NV\\
kazem.taghva@unlv.edu }
\date{} 
\begin{document}
\thispagestyle{empty}
\maketitle
\providecommand{\keywords}[1]{\textbf{\textit{Key Words: }} #1}

\begin{abstract}

Displaying a document in Middle Eastern languages requires contextual
analysis due to different presentational forms for each character of
the alphabet. The words of the document will be formed by the joining
of the correct positional glyphs representing corresponding
presentational forms of the characters. A set of rules defines the
joining of the glyphs. As usual, these rules vary from language to
language and are subject to interpretation by the software developers.

In this paper, we propose a machine learning approach for contextual
analysis based on the first order Hidden Markov Model. We will
design and build a model for the Farsi language to exhibit this
technology. The Farsi model achieves 94 \% accuracy with the training
based on a short list of 89 Farsi vocabularies consisting of 2780
Farsi characters. 

The experiment can be easily extended to many languages including
Arabic, Urdu, and Sindhi. Furthermore, the advantage of this approach
is that the same software can be used to perform contextual analysis
without coding complex rules for each specific language. Of particular
interest is that the languages with fewer speakers can have greater
representation on the web, since they are typically ignored by
software developers due to lack of financial incentives.

\end{abstract}


\section{Introduction}

One of the main objectives of the Unicode is to provide a setting that
non-English documents can be easily created and displayed on modern
electronic devices such as laptops and cellular phones. Consequently,
this encoding has led to development of many software tools for text
editing, font design, storage, and management of data in foreign
languages. For commercial reasons, the languages with high
speaking populations and large economies have enjoyed much more rapid
advancement in Unicode based technologies. On the other hand, less
spoken languages such as Pushtu is barely given attention. According
to \cite{Penzl}, approximately 40 to 60 million people
speak Pushtu worldwide.

Many Unicode based technologies are based on proprietary and patented
methods and thus are not available to the general open source software
developers' communities. For example, BIT \cite{BIT} does not reveal its
contextual analysis algorithm for Farsi\cite{kourosh}. Many software
engineers need to redevelop new methods to implement tools to mimic
these commercial technologies. The new contextual analysis for Farsi
developed by Moshfeghi in Iran Telecommunication Research Center is an
example of these kinds of efforts\cite{kourosh}.

The Unicode also introduces a challenge for the internationalization
of any software regardless of being commercial or open source. Tim
Bray \cite{Bray} writes:

\begin{quotation}
`` Whether you're doing business or academic research or public service,
 you have to deal with people, and these days, it's quite likely that
 some of the people you want to deal with come from somewhere else,
 and you'll sometimes want to deal with them in their own
 language. And if your software is unable to collect, store, and
 display a name, an address, or a part description in Chinese,
 Bengali, or Greek, there's a good chance that this could become very
 painful very quickly.

There are a few organizations that as a matter of principle operate in
one language only (The US Department of Defense, the Académie
française) but as a proportion of the world, they shrink every year.''

\end{quotation}

This internationalization is a costly effort and subject to
availability of resources. As mentioned above, languages with high
speaking population such as Mandarin attract a lot of the
efforts. The availability of data in Unicode represents an opportunity
to employ machine learning techniques to advance software
internationalization and foreign text manipulation. The language
translation technologies heavily use Hidden Markov Models to improve
translation accuracy \cite{Brown:1993:MSM:972470.972474} \cite{Botha2014}. 

In this paper, we propose the use of HMM for contextual analysis. In
particular, we design and build a generic HMM for Farsi that can be
easily adapted to other Middle Eastern languages.

In section~\ref{sec:background}, we provide some background and
related work on contextual analysis. Section~\ref{sec:hmm} will provide
a brief introduction to first order HMM. In section~\ref{sec:farsi},
we describe the design and implementation of our HMM for Farsi
contextual analysis. The training and testing of HMM will be explained
in section~\ref{sec:test}. Finally, section~\ref{sec:conclusion}
describes our conclusion and proposes future work.

\section{Background}
\label{sec:background}

In 2002, the Center for Intelligent Information Retrieval at the
University of Massachusetts, Amherst, held a workshop on Challenges in
Information Retrieval and Language Model \cite{Allan}. The premise of this
workshop was to promote the use of the Language Model technology for
various natural languages. The aim is to use the same software for
indexing and retrieval regardless of the language. It was pointed out
that, by using training materials such as document collections, we can
automatically build retrieval engines for all languages. This report
was one of the reasons that we decided to start a couple of projects
on Farsi and Arabic \cite{itccTaghvaCPN04} \cite{SdiutTaghva03}.

Consequently, these projects led to developments of the two widely
used Farsi and Arabic Stemmers
\cite{ittcTaghvaBS05}\cite{itccTaghvaEC05}. One of the difficulties we
had was the lack of technologies for input and display of Farsi and
Arabic documents\cite{SdiutTaghva03}. For example, we needed an input/display
method that would allow us to enter Farsi query words in a Latin-based
operating system without any special software or hardware. It was
further necessary to have a standard character encoding for text
representation and searching. At the time, we developed a system that
provides the following capabilities:

\begin{itemize} 
\item a web-browser based keyboard applet for input
\item if the web-browser has the ability to process and display Unicode content,
 it will be used
\item if the browser cannot display Unicode content, an auxiliary
process will be invoked to render the Unicode content into
a portable bitmap image with associated HTML to display the image
in the browser.
\end{itemize}

Another area of difficulty that we encountered is that the presence of
white space used to separate words in the document is dependent on the
display geometry of the glyphs. Since Farsi and Arabic are written
using a cursive form, each character can have up to four different
display glyphs. These glyphs represent the four different presentation
forms:

\begin{quotation}
\begin{tabbing}
{\it isolated}: \= \kill
{\it isolated}:\> the standalone character\\
{\it initial}:\> the character at the beginning of a word\\ 
{\it medial}:\> the character in the middle of a word\\
{\it final}:\>  the character at the end of a word
\end{tabbing}
\end{quotation}

We found that depending on the amount of trailing white space following a
final form glyph, a space character may or may not be found in the text. 
This situation came to light when our subject matter experts were developing
our test queries. We found that since the glyphs used to display the final
form of characters had very little trailing space, they were manually adding
space characters to improve the look of the displayed queries.

\subsection{Keyboard Applet}
\label{sec:applet}

The keyboard applet was written in java script. The applet displays a
Farsi keyboard image with the ability to enter characters from both
the keyboard and mouse. The applet also handles character display
conversion and joining of the input data.

\begin{figure*}
\begin{center}
\includegraphics[natwidth=700,natheight=246]{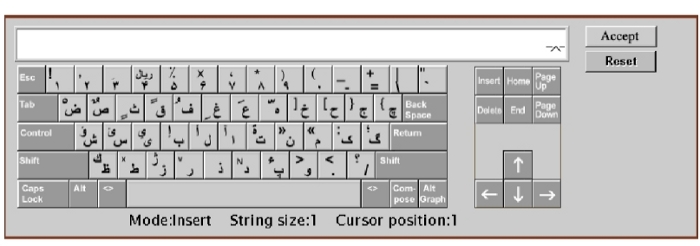}
\caption{Example use of the keyboard applet}
\label{fig:keyboard}
\end{center}
\end{figure*}

The keyboard layout is based on the ISIRI 2901:1994 standard layout as
documented in an email by
Pournader~\cite{Pournader}. Figure~\ref{fig:keyboard} shows the
keyboard applet being used to define our test queries for search and retrieval.

Display of the input data is normally performed by using the preloaded
glyph images. However, if a character has not been preloaded, it can
be generated on the fly. Most of the time, these generated characters
are ``compound'' characters. Farsi (and other Arabic script languages)
may use "compound" characters which are a combination of two or more
separate characters. For example, the rightmost character of
\RL{_hormA}, the Farsi word for ``date'' (that is,
the fruit), is a combination of a kh with a damma.

The complications associated with our work on Farsi and Arabic
convinced us that we need to develop generic machine learning tools if
we want to develop display and search technologies for most of the
Middle Eastern languages. In the next few sections, we will offer a
solution to contextual analysis to display the correct presentational
forms of characters.

\section{Hidden Markov Model}
\label{sec:hmm}

An HMM is a finite state automaton with probabilistic transitions and
symbol emissions
\cite{Rabiner89atutorial}\cite{Rabiner:1990:THM:108235.108253}. An HMM
consists of:
\begin{itemize}

\item A set of states $S = \{s_1, \cdots s_n\}$.

\item An emission vocabulary $V = \{w_1 \cdots w_n\}$.
  
\item Probability distributions over emission symbols where the
  probability that a state $s$ emits symbol $w$ is given by $P(w|s)$. This is denoted by matrix B.
  
\item Probability distributions over the set of possible outgoing
  transitions. The probability of moving from state $s_i$ to $s_j$ is
  given by $P(s_j | s_i)$. This is denoted by matrix A.
  
\item A subset of the states that are considered start states, and
  to each of these is associated an ``initial'' probability that the
  state will be a start state. This is denoted by $\Pi$.
\end{itemize}

As an example, consider the widely used HMM
\cite{DBLP:journals/corr/abs-1212-3817} that decodes weather states
based on a friend's activities. Assume there are only two states of
weather: {\tt Sunny}, {\tt Rainy}. Also assume there are only three
activities: {\tt Walking}, {\tt Shopping}, {\tt Cleaning}.

\begin{figure}[H]
\begin{center}
\includegraphics[natwidth=320, natheight=142]{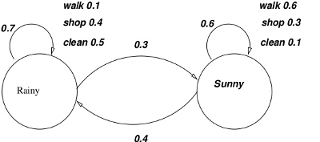}
\caption{An HMM for Activities and weather}
\end{center}
\label{fig:hmm-example}
\end{figure}

You regularly call your friend who lives in another city to find out
about his activity and the weather status. He may respond by saying
``I am cleaning and it is rainy'', or ``I am shopping and it is
sunny''. If you collect a good number of these weather states and
activities, you then can summarize your data as the HMM shown in
Figure ~\ref{fig:hmm-example}. 

This HMM states that on rainy days, your friend walks 10\% of the days
while on sunny days, he walks 60\%. The statistics associated with
this HMM is obtained by simply counting the activities on rainy and
sunny days. 

You also notice arrows from states to states that keeps track of
weather changes. For example, our HMM reflects the fact that
on a rainy day, there is a 70\% chance of rain next day while 30\%
chance of sunshine. 

In addition, one can keep track of how many days in the data are sunny
or rainy. This will be the initial probabilities.  Formally these
statistics are calculated by Maximum Likelihood Estimates
(MLE). Formally, transition probabilities were estimated as:

\begin{equation}
P(s_i,s_j) = \frac{\mbox{Number of transitions from} \ s_i \ \mbox{to}
  \ s_j}{\mbox{Total number
of transitions out of} \ s_i}
\end{equation}

The emission probabilities are estimated with Maximum Likelihood
supplemented by {\em smoothing}.  Smoothing is required because
Maximum Likelihood estimation will sometimes assign a zero probability
to unseen emission-state combinations.

Prior to smoothing, emission probabilities are estimated by:
\begin{equation}
P(w|s)_{ml} = \frac{\mbox{Number of times} \ w \ \mbox{is
    emitted at } \ s}{\mbox{Total number of symbols emitted by
   } \ s}
\end{equation}

The most interesting part of an HMM is the decoding aspect. We may be
told that our friend's activities for the last four days were {\tt
  cleaning, cleaning, shopping, cleaning} and we want to know what the
weather patterns were for those four days. This essentially translate
to finding a sequence of four states $s_1 s_2 s_3 s_4$ that maximizes
p($s_1 s_2 s_3 s_4 | $ {\tt cleaning cleaning shopping
  cleaning}). This amounts to choosing the highest probability among
16 choices for $s_1 s_2 s_3 s_4$. This is computationally very
expensive as the number of states and symbols increases. The solution
is given by the Viterbi algorithm that finds an optimal path using
dynamic programming \cite{Rabiner:1990:THM:108235.108253}. The
algorithm ~\ref{alg:viterbi} is a modification of the pseudo code from
\cite{DBLP:journals/corr/abs-1212-3817}.

\begin{algorithm}[t]
\KwData{ Given K states and M vocabularies, and a
  sequence of vocabularies $ Y = w_1 w_2 \dots w_{n-1} w_n $ }
\KwResult{The most likely state sequence $ R = r_1 r_2 \cdots r_{n-1} r_n $ that maximizes the above probability}

{\bf Function Viterbi}{(V, S, $\Pi$, Y, A, B) : X}\\
\For{each state $s_i$}{
        $T_1[i,1]$ = $\Pi_i * B_{iw_1}$;\\
        $T_2[i,1]$ = $0$;\\
        
    }

\For{ $i = 2,3, \dots , n $ }{
\For{ each state $s_j$ do}{
        $T_1[j,i]$ = $max_k (T_1[k, i-1] * A_{kj} * B_{jw_i})$;\\
        $T_2[j,i]$ = $argmax_k (T_1[k, i-1] * A_{kj} * B_{jw_i})$;\\
            }
}

$z_n = argmax_k (T_1[k,n])$\\
$r_n  = s_{Z_n}$\\

\For{$i = n, n-1, \dots, 2$}{
        $z_{i-1} = T_2[z_i,1]$; \\
        $r_{i-1} = S_{z_i -1}$;
        
}

Return {~R}

\caption{Viterbi Algorithm}
\label{alg:viterbi}
\end{algorithm}
 
In the next section we will describe the design and implementation of
an HMM for Farsi contextual analysis.\\

\section{Farsi Hidden Markov Model}
\label{sec:farsi}

The Farsi HMM is very similar to the example of HMM described in the
previous section. The HMM has a state for each presentation form of
Farsi alphabet. Also the HMM has a vocabulary of size 32, one for each
character in Farsi alphabet.  A simple calculation reveals that the
Farsi HMM should have 128 states and 32 vocabulary.  The HMM has fewer
than 128 states since some of the characters do not have 4
presentational form. For example, there are only two states for the
character \RL{A}, as there are no medial or initial form for this
character. 

\begin{figure}[H]
\begin{center}
\includegraphics[natwidth=229,natheight=40]{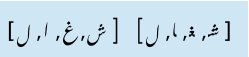}
\caption{The four isolated characters on the left are vocabularies
  while the four characters on the right are states of the HMM}
\end{center}
\label{fig:sequence}
\end{figure}

As an example, suppose we want to type the word \RL{^s.gAl}, in
English {\tt jackal}. On the keyboard, we type four isolated
characters \RL{^s}, \RL{.g}, \RL{A}, and \RL{l}. The HMM should decode
these four characters as {\tt initial}, {\tt medial}, {\tt final}, and
{\tt isolated}, respectively.  In other words, the sequence of the
four isolated characters (or vocabulary in HMM terminology) should be
decoded in the four states as shown in Figure ~\ref{fig:sequence}.

\begin{figure*}
\begin{center}
\includegraphics[natwidth=500,natheight=522]{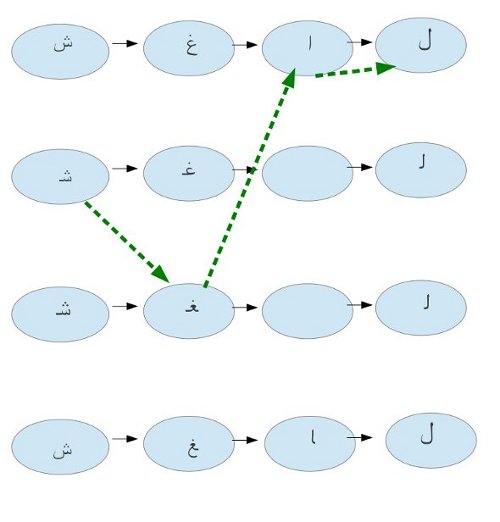}
\caption{Glyphs Chosen by HMM}
\end{center}
\label{fig:farsi-hmm}
\end{figure*}

The part of the HMM as displayed in Figure.~\ref{fig:farsi-hmm} shows
how Viterbi algorithm takes the path to decode the correct
form of the characters by choosing the appropriate states. As we
observe, there are four states for the character \RL{.g} representing
the four shapes of this character. We also observe that there are only
two states for the character \RL{A}, as there are no medial or initial
form for this character. 

A typical implementation of HMM adds states and vocabularies as being
trained\cite{rubyHMM}. The training is done by providing pairs of the form
$([w_1w_2 \dots w_{n-1}w_n], [s_1s_2 \dots s_{n-1}s_n])$ similar to
the $[vocabularies, states]$ sequences as shown in Figure
~\ref{fig:sequence}.

\section{Training and Testing of HMM}
\label{sec:test}

\begin{figure*}
\begin{center}
\includegraphics[natwidth=454,natheight=537]{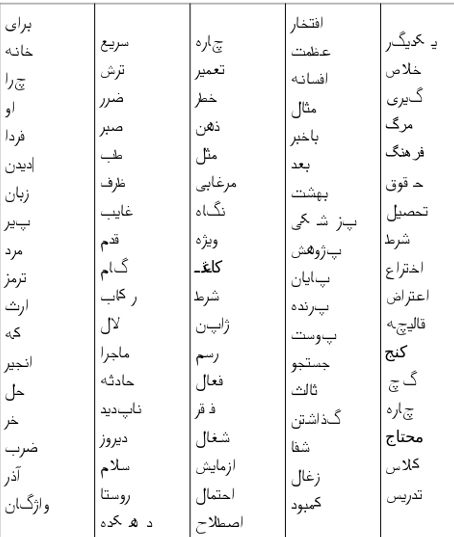}
\caption{Top 89 frequent words from Kayhan newspaper published in 2005}
\end{center}
\label{fig:training-data}
\end{figure*}

We trained the HMM with 89 words ( 2780 characters ) chosen from the
list of the frequent words from Kayhan newspaper published in 2005
\cite{Kayhan2005}. There are over 10,000 words in this collection. We limited
the training to this short list to save time. The list of these words
are shown in Figure ~\ref{fig:training-data}.

\begin{figure*}
\begin{center}
\includegraphics[natwidth=744,natheight= 270]{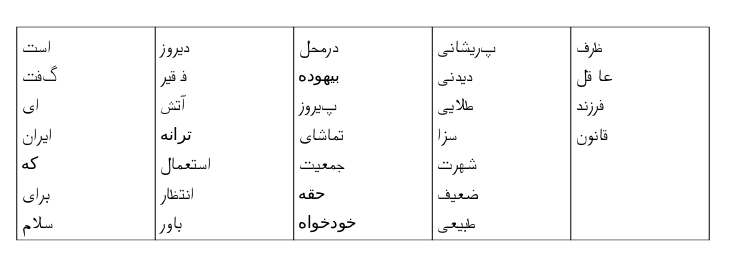}
\caption{Test data chosen randomly}
\end{center}
\label{fig:test-data}
\end{figure*}

The test data is a small number of words selected randomly from a
small dictionary and shown in Figure ~\ref{fig:test-data}. This list
contains 32 words ( 350 characters ). The training file contains pairs
of words separated by a vertical bar. The first word is the isolated form
and the second word is the correct presentational form of the word. We
read the file one line at a time and submit the two words for training
as seen in the following Ruby code:

\begin{quotation}
\begin{tabbing}
f = File.open("./training-data")\\
farsi.train([" "],[" "])\\
f.each do | line |\\
f.each\= \kill
\>seq1,seq2 = line.chomp.split(/s*|s*/)\\
\>farsi.train(seq1.split(" "), seq2.split(" "))\\
end\\
\end{tabbing}
\end{quotation}

As it is seen, we have added a blank vocabulary and state to our
HMM. The HMM adds vocabulary and states as a part of the training. The
HMM has 32 vocabulary and 74 states. It is anticipated that the HMM
will have more states as the size of the training data increases.

The test correctly decoded 94\% of all the characters. Most of the
mistakes are due to the fact that the HMM has not seen enough
combination examples of characters. For example, in the word
\RL{at^s}, the initial form of \RL{t} was not decoded correctly. A
closer examination of the training data reveals that there are no
occurrence of \RL{t^s} in the set. Similarly, there are other errors of
this form such as the initial form of \RL{n} in the word
\RL{traanh}. There are also a few errors attributed to the double
combination of the character \RL{y} as in \RL{.tlaayy}. We believe most
of these errors will be corrected with a larger training sets.

\section{Conclusion and Future Work}
\label{sec:conclusion}

In this paper, We have presented a machine learning approach to the
contextual analysis of script languages. It is shown that an ergodic
HMM can be easily trained to automatically decode presentational forms
of the script languages.

Although the paper is developed based on Farsi, it can be easily
extended to other middle eastern languages. Further training and
research in this area can improve the character accuracy. A successful
program for contextual analysis may have to include a list of
exceptional words that do not fall into the normal combination of the
characters. It is also important to notice that most of the Arabic and
Farsi type setting technologies such as ArabTex \cite{arabtex} or
FarsiTex \cite{farsitex} have problems with contextual analysis. This
is mainly due to the fact that it is practically impossible to devise
an algorithm that has 100\% accuracy for tasks associated with natural
languages.

Finally, a higher order HMM may also improve the contextual
analysis. For example, it is shown that the second order HMM improves
the hand written character recognition \cite{HE}. It may also worth
mentioning that the second order HMM does not improve error detection
and correction for post processing of printed documents \cite{Taghva2013}.

\bibliographystyle{plain}
\bibliography{unicode}

\end{document}